# INVESTIGATING OUTPUT ACCURACY FOR A DISCRETE EVENT SIMULATION MODEL AND AN AGENT BASED SIMULATION MODEL

Mazlina Abdul Majid

Uwe Aickelin

Peer-Olaf Siebers

School of Computer Science
Nottingham University
Nottingham, NG8 1BB, U.K.

School of Computer Science
Nottingham University
Nottingham, NG8 1BB, U.K.

School of Computer Science
Nottingham University
Nottingham, NG8 1BB, U.K.

**ABSTRACT**

In this paper, we investigate output accuracy for a Discrete Event Simulation (DES) model and Agent Based Simulation (ABS) model. The purpose of this investigation is to find out which of these simulation techniques is the best one for modelling human reactive behaviour in the retail sector. In order to study the output accuracy in both models, we have carried out a validation experiment in which we compared the results from our simulation models to the performance of a real system. Our experiment was carried out using a large UK department store as a case study. We had to determine an efficient implementation of management policy in the store's fitting room using DES and ABS. Overall, we have found that both simulation models were a good representation of the real system when modelling human reactive behaviour.

## 1 INTRODUCTION

Simulation has become a preferred tool in Operation Research (OR) for modelling complex systems. Studies in human behaviour modelling have received increased focus and attention from simulation research in the UK (Robinson 2004). The research in human behaviour modelling has been applied to various application areas such as manufacturing, health care, military and many more. As found in the literature, researchers choose either Discrete Event Simulation (DES) or Agent Based Simulation (ABS) as tools to investigate human behaviour problems. The choice of which simulation technique to be used relies on the individual judgment of the simulation model characteristics and their experience with the model. The representation of human behaviour contains complexity and variability; therefore when investigating such systems it is very important to choose a suitable modelling and simulation technique.

In this research, we aim to provide an empirical study in order to find out which simulation modelling technique is a good representation of a real system in our validation experiment. In the validation experiment we have compared the results from traditional DES and ABS models to the performance of the real system. The main difference between traditional DES and ABS is that in the first one the modelling focuses on the process flow, while in the ABS the modelling focus is on the individual entities in the system and their interactions.

Human reactive behaviour means "how a certain individual responds to a certain request. For example, sales staff provides help when needed. In this work, we investigate the output accuracy of DES and ABS models when modelling human reactive behaviour in a department store. Statistical tests were used to compare the models.

The content of the report is as follows: a background section which gives a taxonomy of simulation techniques and a discussion of previous related work. Background in Section 2 explores the theory and characteristics of three major OR simulation methods, i.e. DES, ABS and System Dynamics (SD). In Section 3, we define our case study and the model design. Validation experimentation is presented in Section 4, where we also compare our simulation models' output against the real world output by using quantitative methods. In addition results from the experiment are also discussed. Finally in Section 5 we draw our conclusions and summarize the current progress of our research.

## 2 BACKGROUND

There are several tools and techniques that can be used to model a system. Modelling is a process of abstracting a real world problem into modelling tools. Over the last three decades, simulation has become a frequently used modelling tool in OR (Kelton 2007).

A simulation can be defined as a process of executing a model over time. Its ability to model complex systems has made simulation the preferred user choice when compared to mathematical models. Simulation can be classified into three types; Discrete Event Simulation (DES), System Dynamics (SD) and Agent Based Simulation (ABS).

DES models represent a system based on a series of chronological sequencing of events where each event changes state in discrete time. Meanwhile, SD models represent real world phenomenon using stock and flow diagrams, causal loop diagrams (to represent a number of interacting feedback loops) and differential equations. In contrast to the DES and SD models, ABS models comprise of a number of autonomous, responsive and interactive agents that will interact with each other in order to achieve their objectives. We can summarise the three simulation techniques as follows: the DES and ABS models are suitable to work with discrete event. They both model changes in discrete time from one event to another. On the other hand, SD models are suitable for modelling a system with continuous state changes.

We have found literature performing comparisons between the DES and ABS models together with SD models in terms of model characteristics; however, none of them are currently focusing on modelling human behaviour. One of the relevant papers comparing simulation technique with regards to model characteristics is (Wakeland et al. 2005), where they have compared SD and ABS in field of Biomedical. The authors found that the understanding of the aggregate behaviour in SD models and state changes in individual entities in ABS models is relevant in the study. SD and DES comparisons in the field of Fisheries were described by (Morecroft and Robinson 2006). They found that SD and DES are different approaches but both are suitable for modelling systems over time. Existing comparisons between DES and ABS were described by (Becker et al. 2006) in the field of transportation where they found that DES is less flexible than ABS, i.e. it is difficult to model different behaviours of shippers in DES. In addition to the existing comparison between DES and ABS, (Yu et al. 2007) performed a quantitative comparison on the DES and ABS models outputs in the field of transportation. We found just one work that describes the three techniques discussed in this study. (Owen et al 2008) tried to establishing a framework for comparing the different modelling techniques.

We have also found a disparity in the quantity of work comparing SD and ABS or SD and DES contrasted to the amount of work comparing DES and ABS. Specifically, this disparity can be outstanded when referring to modelling human behavioural and where the focus is output accuracy. Therefore, in this research we choose to study the differences between DES and ABS in this regard.

To study the differences in both models, we choose to focus on management practices in retail with regards to worker behaviour. Research in retail previously focused on consumer behaviour (i.e. Schenk et al. 2007). However, research in management practices has started to evolve as described by (Siebers et al. 2007, 2008). As discussed in the existing comparison of simulation techniques above, a lot of work has been done to compare simulation techniques in transportation and supply chain management. These studies have focused on modelling characteristics. On the other hand, we have decided to compare the accuracy of output in the DES and ABS models on management practices, as it is currently a developing area of study in the retail domain.

## 3 CASE STUDY FIELDWORK

In order to achieve our aim, we have used a case study approach. The research has focused on the operation of the main fitting room in the womenswear department of one of the top ten retailers in the UK (see Figure 1). We wanted to identify the potential impact for fitting room performance when having different numbers of sales staff permanently present.

We have investigated the staff behaviour on human reactive behaviour, which relates to staff responding to the customer when being available and requested. From the case study exploration, we have produced a flow chart diagram for DES conceptual models (see Figure 2) as in DES we focus on process flow. Our ABS conceptual models (see Figure 3 for an example) are state chart diagrams for the different types of people we had to represent (in our case customers and staff) as in ABS we focus on the individual 'actors' and their interactions.

In the fitting room operation, the staff reactive behaviours can be seen in three jobs. Job 1 is about counting the number of garments and giving the green card, job 2 is about providing help and lastly job 3 is about receiving the green card and unwanted garments from the customers. The case study data has been transformed into the simulation inputs for the experimentation.

## 4 EXPERIMENTATION

Two similar simulation models of DES and ABS were developed using multi-paradigm software known as Anylogic$^{tm}$. A conventional M/M/1 queuing system is constructed in both techniques. The model consists of an arrival process (customers), three single queues (customer entry queue, customer return queue, customer help queue) and resource (sales staff). The run length for both simulation models is one day from 9.00 am to 5.00 pm and was replicated to 100 runs. Both simulation models used the same inputs. The simulation model we developed has one member of staff that does all three jobs mentioned above, with a workload of 45% for job 1 (counting garments on entry), a workload of 10% for job 2 (providing help) and a workload of 45% for job 3 (counting garments on exit). Along with developing the simulation models for DES and ABS, the verification and validation process is performed simultaneously with the models. In the next sub-section,

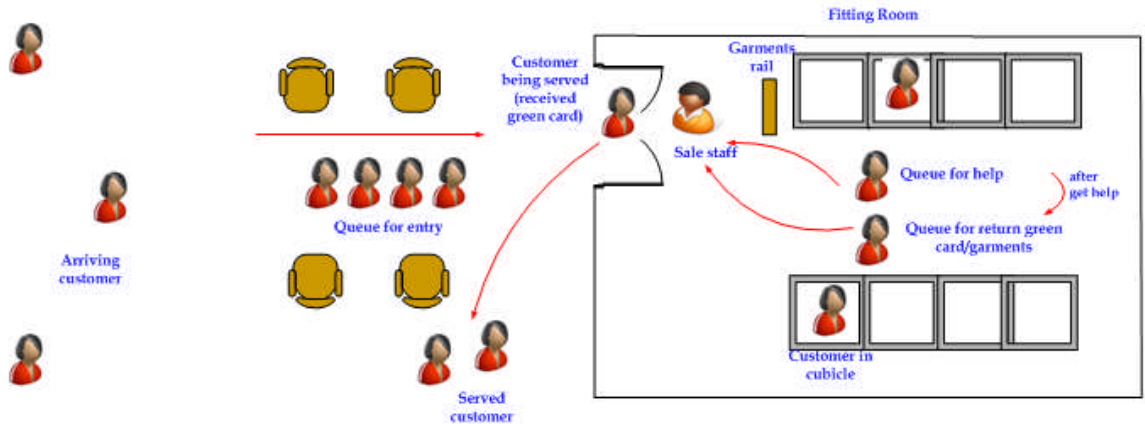

Figure 1: An illustration of the main fitting room operation

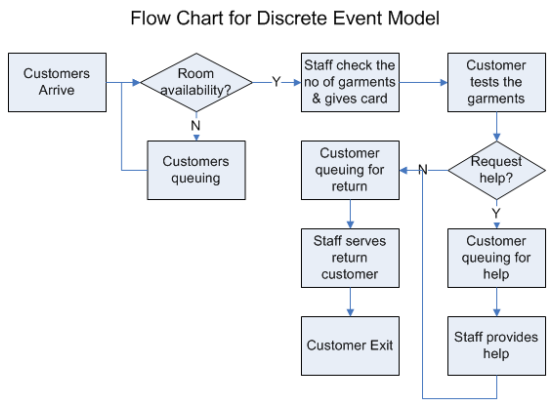

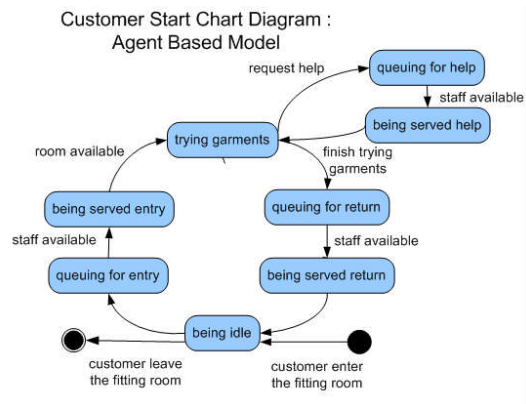

Figure 2: Flow chart for DES model                Figure 3: State chart for ABS model (customer)

we describe how we have performed our validation experiment.

### 4.1 Validation Experiment

We have used black box validation to compare the simulation outputs from DES and ABS with the real system output using quantitative method. By using statistical methods for this comparison, we are able to find out which simulation model is a good representation of a real system. We compared output data observed in the department store to the distribution of the predicted output generated by each model. By assuming the alternative hypothesis is the opposite of null hypothesis in all tests, thus in the paper we will only state the null hypothesis. The main hypothesis for the validation test was constructed as following:

$H_o$ = DES model is a good representation of a real system
$H_o$ = ABS model is a good representation of a real system

We have used *mean waiting time* from the three queues as our performance measurement in the experiment. This was the only performance data we were able to collect from the real system. Two tests have been setup in answering the hypothesis. These tests are described in the following sub-section.

#### 4.1.1 Test 1: Comparing Medians Using a Non Parametric Test

In Normal distributed data, the observed mean, median and mode will have similar values because they are identical in the population. Since waiting time data are not normally distributed, we have chosen to compare medians as a more robust measure of central tendency rather than means and have chosen a Non Parametric statistical test known as Mann-Whitney to avoid assuming Normality. We have constructed the following hypothesis for the test:

*Hₒ = DES models are not significantly different with real system in mean customer waiting time*

*Hₒ = ABS models are not significantly different with real system in mean customer waiting time*

In performing the Mann-Whitney test, we have used the open source statistical software, R. The median of waiting time from DES and ABS models and real system were used for the comparison purpose. We have chosen 0.05 as the level of significance. If the probability of seeing data as or more different than expected under the model is smaller than 0.05, we reject the null hypothesis. If it is not then we fail to reject the null hypothesis. Put some simply we can conclude that the data we observed are consistent with the model's predictions (without assuming that the model must itself therefore be correct). In the statistic test for DES model against real system, p-value is 0.3269. Since the p-value is not less than the chosen sigma value of 0.05; there is insufficient evidence to reject our hypothesis. The statistic test for ABS model against real system showed similar results when it had a p-value of 0.2958, and it was larger than the chosen level of 0.05. Therefore, we fail to reject our hypothesis for both simulation models. We find the distribution of median waiting times is consistent with both models. Next we look at the variability of waiting time to see if the one we get from our simulation models matches the variability we can observe in the real system.

### 4.1.2 Test 2: Measuring Variability

In finding out which DES and ABS models are having more variation against real system we have plotted a frequency distribution of customer waiting time for a single day (represents one replication) from both simulation models and real system (Figure 4).

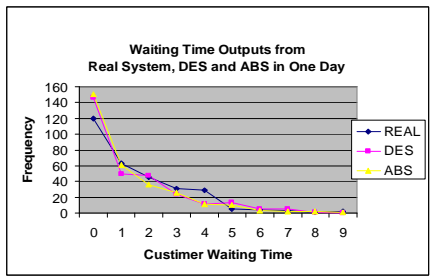

Figure 4: Waiting time outputs for one day in real system, DES and ABS models.

Next, we calculate the variance (a measure of dispersion) of customer waiting times. This allows us to compare the variability of the simulation outputs data against real system on a statistical basis by study the spread of frequency distribution of customer waiting time from both simulation models and real system. To execute the test, we have produced the following hypothesis:

*Hₒ = DES model shows similar variability compare to the real system*

*Hₒ = ABS model shows similar variability compare to the real system*

The spread of frequency distribution for DES and ABS models (as shown in Figure 4) seem slightly different and to look into their differences, we have compared the variances of the outputs. The variance shows how close the simulation outputs i.e. waiting time, in the distribution are to the middle of the distribution.

Table 1: Waiting time outputs for DES, ABS and real system in one day with standard deviation and variance.

| Models | Mean waiting time (minutes) | Standard deviation | Variance |
|---|---|---|---|
| Real system | 1.68 | 1.73 | 3.01 |
| DES | 1.69 | 1.59 | 1.96 |
| ABS | 1.61 | 1.70 | 2.89 |

By looking at the result in Table 1 above, the variance between DES model and the real system is significantly different with 35% difference and ABS model is similar with real system with 4% difference. The reason for the differences in variance for both simulation models may be due to the different operation structure in DES and ABS models. The entities in DES model are having a strong order and they are depending to each other to change their state. Meanwhile agents in ABS model are in decentralised order and they change their state independently. For this test we fail to reject our hypothesis for ABS model where the model has produced similar variability with the real system. However, we have to reject the hypothesis for DES model where the model has shown dissimilarity in variability when compared to the real system.

### 4.1.3 Validation Experiment Conclusion

In conclusion we find on statistically significant differences between the DES and ABS models when compared to the real system in Test 1. However in Test 2 we found DES model produced different variability then the real system and ABS produced the similar variability with the real system. Therefore based on Test 2 result, we suggest that the ABS model is more suitable in representing the behaviour of the real system where the operation involving human is the main focus of the system. Nevertheless based on the validation experiment both simulation models are good in representing the real system and ABS shows better representation of the real system when passed the Test 2.

## 5   CONCLUSION AND FUTURE WORK

In the validation experiments between DES and ABS models, we are able to demonstrate which simulation models is a good representation of real system when modelling human reactive behaviour. To achieve that, we compared the output accuracy of the simulation outputs with the real system in mean customer waiting time. Statistical methods to analyse the outputs have helped us to identify the statistical significance of the similarity and difference of the simulation models to one another. The testing suggested that even though both DES and ABS models produce similar outputs in Test 1, when we compared their medians to the real system, they showed different variation in their model outputs in Test 2. ABS models reflect the real system behaviour much better than DES model in terms of their predicted variability in waiting times. The system that we modelled in DES and ABS is a typical queuing system with no extra features of complex human behaviour. Moreover, based on the validation results we have concluded that both DES and ABS models are a good representation of the real system that contains human reactive behaviour. Investigating the outputs' behaviour using different sample data and different scenarios will be carried out as part of our future work. We also want to evolve our experiment by modelling human proactive behaviour in ABS only and look into the benefit of modelling such behaviour in a queuing system for the OR simulation study.

## AUTHORS BIOGRAPHIES

**MAZLINA ABDUL MAJID** is a PhD student in the School of Computer Science, University of Nottingham. Her interest is in discrete event simulation and agent based simulation. Her email is<mva@cs.nott.ac.uk>.

**Prof. UWE AICKELIN** is a Professor in the School of Computer Science, University of Nottingham. His interests include agent-based simulation, heuristic optimization and artificial immune system His email is<uxa@cs.nott.ac.uk>.

**Dr. PEER –OLAF SIEBERS** is a Research Fellow in the School of Computer Science, University of Nottingham. His main interest includes agent-based simulation and human oriented complex adaptive system. His email is <pos@ cs.nott.ac.uk>.